\documentclass{ceurart}
\usepackage{latexsym}
\usepackage{amssymb}
\usepackage{amsmath}
\usepackage{amsthm}
\usepackage{booktabs}
\usepackage{enumitem}
\usepackage{graphicx}
\usepackage{color}
\usepackage{algorithm}
\usepackage{algpseudocode}

\newcommand{\BibTeX}{B\kern-.05em{\sc i\kern-.025em b}\kern-.08em\TeX}
\usepackage{listings}
\begin{document}

\copyrightyear{2024}
\copyrightclause{Copyright for this paper by its authors.
  Use permitted under Creative Commons License Attribution 4.0
  International (CC BY 4.0).}

\conference{EXPLIMED - First Workshop on Explainable Artificial Intelligence for the medical domain - 19-20 October 2024, Santiago de Compostela, Spain}

\title{AI Readiness in Healthcare through Storytelling XAI}



\author[1,2]{Akshat Dubey}[%
orcid=0009-0008-4823-9375,
email=DubeyA@rki.de,
]
\author[1]{Zewen Yang}[%
orcid=0000-0001-9974-3231,
email=ZewenY@rki.de,
]
\author[1,2]{Georges Hattab}[%
orcid=0000-0003-4168-8254,
email= HattabG@rki.de
]

\address[1]{Center for Artificial Intelligence in Public Health Research (ZKI-PH) at Robert Koch Institute, Nordufer 20, Berlin 13353, Germany}
\address[2]{Department of Mathematics and Computer Science, Free University of Berlin, Arnimallee 14, Berlin 14195, Germany}

\maketitle
\begin{abstract}
Artificial Intelligence is rapidly advancing and radically impacting everyday life, driven by the increasing availability of computing power. Despite this trend, the adoption of AI in real-world healthcare is still limited. One of the main reasons is the trustworthiness of AI models and the potential hesitation of domain experts with model predictions. 
Explainable Artificial Intelligence (XAI) techniques aim to address these issues.
However, explainability can mean different things to people with different backgrounds, expertise, and goals. 
To address the target audience with diverse needs, we develop storytelling XAI. 
In this research, we have developed an approach that combines multi-task distillation with interpretability techniques to enable audience-centric explainability. 
Using multi-task distillation allows the model to exploit the relationships between tasks, potentially improving interpretability as each task supports the other leading to an enhanced interpretability from the perspective of a domain expert. 
The distillation process allows us to extend this research to large deep models that are highly complex. 
We focus on both model-agnostic and model-specific methods of interpretability, supported by textual justification of the results in healthcare through our use case. 
Our methods increase the trust of both the domain experts and the machine learning experts to enable a responsible AI.

\end{abstract}



\section{Introduction}

The adoption of artificial intelligence (AI) in healthcare has been relatively slow compared to other industries, despite its immense potential benefits.
Acceptance of AI by patients and healthcare providers requires building trust in these AI-powered systems in hospitals and healthcare systems~\cite{celi2019awakening,davenport2022factors}.
Explainability techniques from the field of Explainable AI (XAI) aim to increase the transparency of AI decision-making, enabling clinicians and patients to understand and validate AI results, thereby building trust~\cite{reddy2022explainability}.
With explainable AI, medical professionals can share the reasoning behind their decisions, maintaining accountability and empowering patients to make informed decisions about their care~\cite{amann2020explainability}.
While current explainability methods provide good insight into the workings of Machine Learning (ML) and deep learning models, they are not yet suitable for domain experts who have little or no experience with such models. 
Trustworthy AI techniques have gained significant attention spanning various domains~\cite{yang2024whom}.
Explanations need to be tailored to the particular goals, concerns, and decision requirements of the non-specialist audience, highlighting the data that is most important to them~\cite{jin2023invisible}.
\\
\\
There has been previous work in this area, but the research is limited in terms of audience-centric explainability through the utilization and combining currently available datasets for diverse tasks.
The current research methods attempt to explain the models in terms of heat map visualizations or post hoc methods including the Local Interpretable Model-agnostic Explanations (LIME) method or SHapley Additive Explanations (SHAP) ~\cite{example_1, example_2}.  
Motivating works in Storytelling XAI already exist. 
SHAPstories and CFstories, are two new techniques that use large language models to generate narratives explaining AI predictions based on SHAP values and counterfactual explanations~\cite{tell_me_a_story}.
Counterfactual explanations frequently rest on the presumptions that the features are independent and that the changes in the highlighted features match real-world activities. 
In actuality, though, the characteristics are not independent, and actions are likely to have simultaneous effects on several features. 
Post-hoc counterfactual explanations need to provide faithful changes that a human can implement in practice and should conform to the observed correlations in the training data rather than being anomalies. They may not be faithful to the original data due to problems like overfitting or excessive generalization, which results in unsatisfactory interpretability.
Although ground truth plausibility of counterfactuals is acknowledged as a crucial attribute, the search results show that this characteristic is still challenging to measure in practice~\cite{asemota2024longitudinal, baron2023explainable}.
The research also utilizes the application of the Large-Language-Models (LLMs) to enable narrative XAI, another limitation to implementing this framework in healthcare and finance.
The XAINES project~\cite{xaines} also explores the use of narratives to explain AI systems, hypothesizing that narratives are an appropriate means to communicate explanations, particularly to users without machine learning backgrounds.
Whether the intended audience is developers, domain experts, or end users who are affected by the AI's decisions, the project intends to make it possible to explain AI systems in a way that is specific to their requirements and expectations.
An essential component of the XAINES approach is the narratives used to describe the AI, which facilitates the successful communication of the causal chain of events leading to the AI's behavior. 
The challenges of translating causal reasoning into effective explanatory narratives and interactive experiences for diverse audiences still exist.
However, the existing storytelling XAI frameworks do not adequately address the effective usage of available datasets, and how to combine them and achieve end-to-end audience-centric XAI without relying on distinct models trained for specific tasks.
\\
\\
We develop a storytelling XAI framework to provide end-to-end explanations customized for domain specialists, in our case, healthcare practitioners as well as ML practitioners. 
The storytelling XAI uses a knowledge distillation~\cite{hinton2015distilling} approach coupled with multi-task prediction and combines it with interpretability techniques to achieve explainability. This approach helps to achieve concept-based explainability. Concept-based explainability methods aim to explain the behavior and predictions of deep neural networks (NN) using human-understandable concepts such as texts.
The knowledge distillation enables us to utilize the current available datasets, combine them with different tasks, and have a single model to deal with all the different tasks.
To generate explanations, our research uses both model-agnostic and model-specific methods. 
The model-agnostic methods use an approximation for interpretability, which allows flexibility in model selection but limits the accuracy of interpretations due to the approximations involved. 
Whereas the model-specific methods can interpret the results from the components of the model giving us accurate interpretation but limiting the flexibility to choose models~\cite{carrillo2021individual}.
By combining knowledge distillation, multiple datasets for diverse tasks, and interpretability techniques, we achieve explainability that is useful for domain experts and ML practitioners to decipher the decision-making process of the AI models.
For this research, we have considered the use case of chest X-ray analysis to demonstrate the effectiveness of our framework in enabling XAI for healthcare professionals.
This work also aims at an end-to-end application of trustworthy AI in healthcare using multi-task predictions supported by interpretation techniques. 
Through this research, complex deep neural networks could also be used in healthcare without compromising model trustworthiness.
Storytelling XAI framework for the healthcare domain provides concept-based understanding in human terms to healthcare professionals while providing technical interpretation to ML practitioners.




\section{Background}

\subsection{Knowledge Distillation}
Knowledge distillation refers to the process of transferring knowledge from a comparatively large model to a smaller model without compromising on performance. 
The complex large model is called \textit{teacher} and the smaller model is called \textbf{student}. 
The rationale behind model distillation is to train complex large models also called \textit{teacher} for a specific task and then transfer the knowledge to a smaller model also called \textit{student} using distillation loss (Equation: ~\ref{eq:distillation_los}). 
The loss function employed in knowledge distillation to instruct the student model to imitate the teacher model's behavior is called the distillation loss. 
By using a "temperature" scaling function in the softmax, the logits are softened, so smoothing down the probability distribution and exposing the teacher's taught inter-class correlations. The probability \textit{p\textsubscript{i}} of class \textit{i} from the logits \textit{z} is calculated as:
\begin{eqnarray}\label{eq:distillation_los}
p_i = exp(z_i/T) \div \sum_{j} exp(z_j/T)
\end{eqnarray}
where \begin{math}
    T
\end{math} is the parameter of temperature and when 
\begin{math}
    T=1
\end{math} then we get the original softmax function. 
The probability distribution produced by the softmax function softens with increasing T, giving more insight into which classes the teacher thought were more like the ones that were expected.
The discrepancy between the instructor model's soft targets and the student model's predictions is quantified by this loss function. The student model learns the internal representation from the teacher model~\cite{hinton2015distilling}.   
The student model learns not only the target outputs but also the internal representations and similarity information from the teacher model. 
This allows the student model to capture the same high-level concepts and reasoning as the teacher but in a more transparent and interpretable form~\cite{han2023impact, liu2018improving}. 

\subsection{Multi-task Learning}
In AI, multi-task learning is the process of teaching one model to handle several related tasks at once, as opposed to training different models for every task. The main concept is to increase overall learning efficiency and generalization performance by utilizing shared representations and commonalities across tasks
Multi-task Learning can be implemented through various methods and the two most common methods are shared base network extractors with task-specific heads in which the model comprises task-specific output heads after a shared base network that extracts features.
The second method is the shared decision-making layer in which task-specific layers are connected to the shared decision-making layer in the model~\cite{crawshaw2020multi}. 

\subsection{Interpretability Techniques}

Interpretability allows ML practitioners to understand the connections between the features that go into a model and the results it produces, as well as the relative importance of different features in the decision-making process.
Some models, such as linear regression and decision trees, are inherently interpretable due to their simple structure, while complex models, such as deep neural networks, are often considered black boxes and require additional techniques to enhance their interpretability~\cite{ribeiro2016should}.
The two most common types of interpretation are model-agnostic methods and model-specific methods. 
Model-specific methods are designed to interpret specific types of models. 
Visualization techniques and attention mechanisms (Equation:\ref{eqn:attention}) are used to interpret Convolutional Neural Network (CNN) architectures and Recurrent Neural Network (RNN) or transformer architectures, respectively.
\begin{eqnarray}
\label{eqn:attention}
\text{Attention}(Q, K, V) = \text{softmax}\left(\frac{QK^T}{\sqrt{d_k}}\right)V
\end{eqnarray}
where,\\
\begin{math}
    \text{Q :represents the query matrix.}\\
    \text{K :represents the key matrix.}\\
    \text{V :represents the value matrix.}\\
    {d}_{k}\text{ :represents the dimensionality of the key vectors.}\\
\end{math}
This equation computes the attention weights by taking the dot product of the query and key matrices, scaling it by 
\begin{math}
    \sqrt{d}_{k}
\end{math}, applying the softmax function, and then multiplying it by the value matrix to get the attention output.

Visualization techniques include the analysis of saliency maps, and activation maps of the intermediate layers of a model, the most relevant being GradCAM (Equation:\ref{eqn:gradcam}).

\begin{eqnarray}\label{eqn:gradcam}
    L_{\text{Grad-CAM}}^{c} = \text{ReLU}\left(\sum_{k} \alpha_{k}^{c} A^{k}\right)
\end{eqnarray}
where,\\
\begin{math}
    L_{\text{Grad-CAM}}^{c}:\text{represents the Grad-CAM score for class. }{c} \\
    \alpha_{k}^{c} A^{k}: \text{represents the importance of the }{k }\text{th class.}\\
    A_{k}: \text{represents the }{k}\text{th activation map for the last convolution layer.}\\
    \text{ReLU denotes the rectified linear unit function, which clips negative values to zero.}\\
\end{math}

    \begin{figure*}[ht]
    \centering
    \includegraphics[width=\textwidth, angle=0, origin=c]{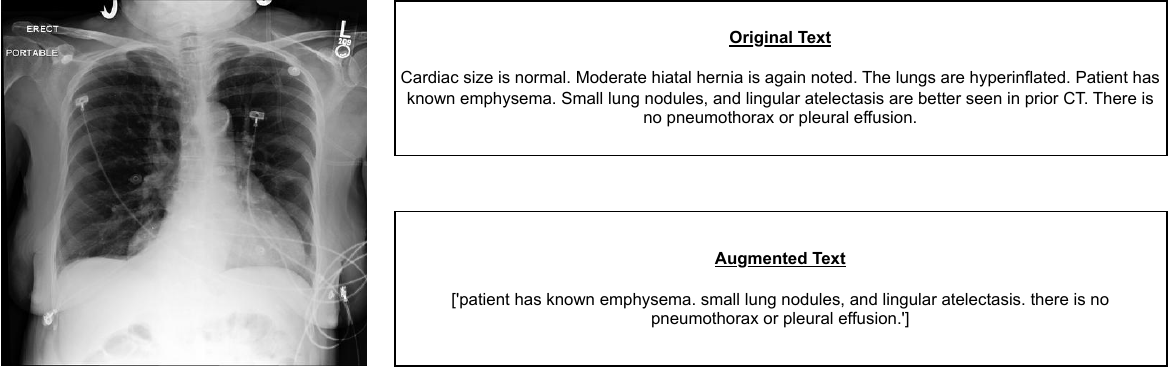}
    \caption{
    A sample chest X-ray with the corresponding report with original text and extended text. 
    Left: X-ray image of a human chest.
    Top right: Original text content describing the prognosis of the chest x-ray in medical jargon.
    Bottom right: Augmented text describing the patient's lung in a concise and clear manner.
    }
    \label{fig:sample_aug}
    \end{figure*}
    
Attention mechanisms are usually implemented to visualize text-based models.
It helps the user to analyze and infer the specific part of the sentence that influences the predictions.
Model agnostic methods are flexible in terms of model choice. 
They are a good choice for interpreting the predictions of a wide variety of models. 
The most common methods related to model-agnostic techniques are Local Interpretable Model-agnostic Explanations (LIME)(Equation:\ref{eqn:lime}) and SHapley Additive exPlanations (SHAP).
\begin{eqnarray}\label{eqn:lime}
    \text{LIME: } \hat{f}_{\text{lime}}(x) = \arg\min_{f \in \mathcal{F}} \mathcal{L}(f, \pi_{x}) + \Omega(f)
\end{eqnarray}
where,\\
\begin{math}
    \hat{f}_{\text{lime}}(x):\text{represents the local interpretable model produced by LIME for input. }{x}\\
    \mathcal{F}: \text{denotes the set of possible interpretable models.}\\
    \mathcal{L}(f, \pi_{x}): \text{represents the loss function measuring how well the interpretable model. }\\
    {f }\text{: approximates the prediction function }\pi_{x}\text{in the local neighborhood of }{x}\text{.}\\
    \Omega(f ):\text{represents a regularization term that encourages simplicity or interpretability of the model }{f}\text{.}\\
\end{math}
LIME approximates the behavior of the model locally using interpretable models such as linear regression by perturbing the input and correspondingly reflecting it in the predictions~\cite{linardatos2020explainable}. 

\section{Storytelling XAI in Healthcare}

Explainable Artificial Intelligence (XAI) is critical in the medical and healthcare domains because AI-driven systems can have serious and even life-threatening consequences for patients if their decisions are not transparent and interpretable. 
Model developers are primarily interested in the performance, stability, and robustness of the AI model. Their focus is on using explainability to debug models and improve accuracy~\cite{hadji2021audience}.
XAI aims to shed light on the "black box" nature of complex AI models such as deep neural networks, which can be difficult for humans to understand. This is important in healthcare to build trust and acceptance of AI systems among clinicians and patients.
Integrating XAI into clinical decision support systems is important to align with core ethical principles in medicine, such as autonomy, beneficence, and non-maleficence~\cite{amann2020explainability}.
There is an increasing need for collaboration between medical and AI experts to develop appropriate frameworks for designing and implementing XAI solutions in the medical domain~\cite{prentzas2023explainable}.
However, explainability in AI is highly dependent on the target audience and their specific needs and knowledge levels. Different stakeholders, such as model developers, business managers, and end users, have different requirements for AI explanations. As models become more complex, the explanations generated by AI techniques can also become more complex and difficult for non-expert users to understand~\cite{rong2023towards}.
Developing audience-specific explanations is challenging as AI systems become more complex. However, this is critical to building trust and acceptance of AI across different stakeholders.
\\

By using Storytelling XAI, the gap between ML engineers and healthcare professionals is reduced. ML engineers can use Storytelling XAI to transform complex ML models into understandable information that can be shared with healthcare domain experts. ML engineers can more easily communicate the model's underlying logic and decision-making processes through storytelling, as opposed to providing raw data or technical model outputs, making it easier for domain experts to understand how the model's predictions fit into actual patient situations and scenarios. Domain experts are more likely to accept the model's suggestions and incorporate AI into clinical decision-making processes if they can follow the reasoning behind the model's outputs through a narrative.

\section{Materials and Methods}
For this research, we have used chest X-ray images to detect the abnormalities in the chest, identify the affected regions, and generate a report generation.
The report generation task generates a report out of the chest X-image and serves as a text justification for the other task.
\subsection{Dataset and Pre-processing}
Our research contains two different tasks which utilize a different dataset for each task. The task and the corresponding dataset used are as follows:

\begin{enumerate}
    
    \item Chest X-ray report generation: We used a pre-processed dataset provided by the authors of MedVill (Medical Vision Language Learner). The pre-processed dataset was prepared from the original MIMIC-CXR-JPG dataset~\cite{johnson2019mimic} and Open-I dataset~\cite{demner2016preparing}. The MIMIC-CXR-JPG dataset contains 377,110 chest X-ray images and corresponding free-text reports. The Open-I dataset contains 3,851 reports and 7,466 Chest X-ray images, consisting of both the lateral and frontal view of the images~\cite{moon2022multi}. For our ease, the lateral views were removed from the dataset. For our experimentation, we have selected 5000 image-report pairs randomly. The reports were augmented with the help of clinical T5 large which (Fig:\ref{fig:sample_aug}) is a sequence-to-sequence transformer (available on HuggingFace) to obtain the summarized version of the radiology report with the sequence length of 128 words~\cite{chizhikova2023sinai}. We performed this step to obtain keywords having higher weightage in the text report through summarization.

    \item Chest X-ray abnormality detection: For this task we used a modified version of the VinDr-CXR dataset~\cite{vinbigdata-chest-xray-abnormalities-detection}. The dataset contains 15,000 samples of DICOM images labeled with 14 abnormalities namely aortic enlargement, atelectasis, calcification, cardiomegaly, consolidation, ILD, Infiltration, Lung Opacity, Nodule/Mass, Other lesion, Pleural effusion, Pleural thickening, Pneumothorax, and Pulmonary fibrosis along with bounding box for detection and localization of the detected anomalies. The dataset was annotated by a group of seventeen expert radiologists~\cite{nguyen2012vindr}. We selected 5,000 samples randomly without unbalancing the class ratio which could have potentially led to the problem of class imbalance.

    \item Lung Segmentation: For this task, we have used the Montgomery dataset which contains images from the Department of Health and Human Services, Montgomery County, Maryland, USA. The dataset consists of 138 CXRs, including 80 normal patients and 58 patients with manifested tuberculosis (TB). The CXR images are 12-bit gray-scale images of dimension 4020 × 4892 or 4892 × 4020. Only the two lung mask annotations are available which were combined into a single image to make it easy for the network to learn the task of segmentation. To make all images of symmetric dimensions we padded the pictures to the maximum dimension in their height or width such that images are 4892 x 4892, this is done to preserve the aspect ratio of CXR while resizing. We scale all images to 512x512 pixels, which retains sufficient visual details for vascular structures.

All the images which were not in JPEG format were converted into JPEG from DICOM format.
The images were normalized to the standard range of [0-1] and resized to 512x512 pixels.

\end{enumerate}

\subsection{Methodology}

\begin{algorithm}
\caption{Student Model Training with Knowledge Distillation For Multi-Task Prediction}
\label{Student_model}
\begin{algorithmic}[1]
\State Initialize student model $S$ with three heads: $S_{report}$, $S_{abnormality}$, $S_{segmentation}$
\State Initialize teacher models $T_{report}$, $T_{abnormality}$, $T_{segmentation}$
\State Initialize loss functions for each task: $L_{report}$, $L_{abnormality}$, $L_{segmentation}$
\State Initialize hyperparameters: learning rate, temperature $T$ and distillation weights $\alpha$
\State Freeze parameters of $S_{abnormality}$ and $S_{segmentation}$
\For{epoch = $1$ to $N$}:
    \For{$batch$ in $training\_data$}
        \State Forward pass through teacher model for report generation: $output_{report} = T_{report}(batch)$
        \State Compute loss for report generation: $L_{report}(S_{report}(batch), output_{report})$
        \State Compute knowledge distillation loss for report generation: $\mathcal{L}_{distillation}(S_{report}(batch), output_{report})$
        \State Compute total loss for report generation: \( \mathcal{L} = (1 - \alpha) \cdot \text{L}_{\text{report}} + \alpha \cdot \mathcal{L}_{\text{distillation}} \)
        \State Backpropagation: Compute gradients \( \nabla_{\theta_s} \mathcal{L} \)
        \State Update parameters: \( \theta_s \gets \theta_s - \eta \cdot \nabla_{\theta_s} \mathcal{L} \)
    \EndFor
\EndFor
    
\State Freeze parameters of $S_{report}$ and $S_{segmentation}$
\For{$epoch = 1$ to $N$}
    \For{$batch$ in $training\_data$}:
        \State Forward pass through teacher model for abnormality detection: $output_{abnormality} = T_{abnormality}(batch)$
        \State Compute loss for abnormality detection: $L_{abnormality}(S_{abnormality}(batch), output_{abnormality})$
        \State Compute knowledge distillation loss for report generation: $\mathcal{L}_{distillation}(S_{abnormality}(batch), output_{abnormality})$
        \State Compute total loss for abnormality detection: \( \mathcal{L} = (1 - \alpha) \cdot \text{L}_{\text{abnormality}} + \alpha \cdot \mathcal{L}_{\text{distillation}} \)
        \State Backpropagation: Compute gradients \( \nabla_{\theta_s} \mathcal{L} \)
        \State Update parameters: \( \theta_s \gets \theta_s - \eta \cdot \nabla_{\theta_s} \mathcal{L} \)
    \EndFor
\EndFor

\State Freeze parameters of $S_{report}$ and $S_{abnormality}$
\For{$epoch = 1$ to $N$}
    \For{$batch$ in $training\_data$}:
        \State Forward pass through teacher model for segmentation: $output_{segmentation} = T_{segmentation}(batch)$
        \State Compute loss for abnormality detection: $L_{segmentation}(S_{segmentation}(batch), output_{segmentation})$
        \State Compute knowledge distillation loss for report generation: $\mathcal{L}_{distillation}(S_{segmentation}(batch), output_{segmentation})$
        \State Compute total loss for segmentation: \( \mathcal{L} = (1 - \alpha) \cdot \text{L}_{\text{segmentation}} + \alpha \cdot \mathcal{L}_{\text{distillation}} \)
        \State Backpropagation: Compute gradients \( \nabla_{\theta_s} \mathcal{L} \)
        \State Update parameters: \( \theta_s \gets \theta_s - \eta \cdot \nabla_{\theta_s} \mathcal{L} \)
    \EndFor
\EndFor
\end{algorithmic}
\end{algorithm}

The storytelling XAI framework (Fig:\ref{fig:framework} is divided into three parts. The priority intent of this framework is not to compete with current state-of-the-art models in terms of evaluation metrics.
This framework utilizes knowledge distillation, interpretability, and datasets for a variety of tasks. Using datasets from different origins allows the framework to generalize better for real-world scenarios as well. 
The three parts involved are:
\begin{figure*}
        \centering
    \includegraphics[scale=0.35]{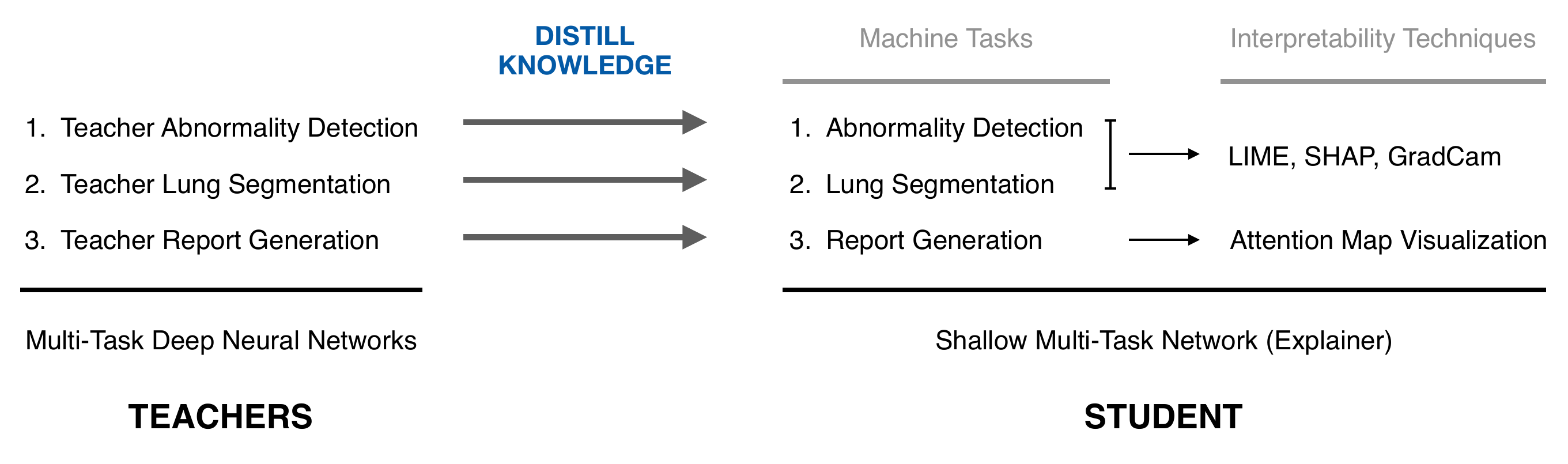}
    \caption{Storytelling XAI Framework. Three teacher networks are trained to perform the specific tasks numbers 1, 2, and 3. Using knowledge distillation, the student model acquires knowledge to perform all the distinct tasks alone. The student model learns the underlying relationship between different features enhancing the interpretability.}
    \label{fig:framework}
\end{figure*}

\begin{enumerate}
    \item The First step involves training the complex deep neural networks for individual tasks. For the task of abnormality detection and localization, a CNN-based model with ResNet 50 backbone is trained using a categorical cross-entropy loss and mean Average Precision(mAP), which also achieves a remarkable performance on this task~\cite{resnet50}. For the task of report generation, a CNN-RNN-Attention model with ResNet50 backbone is trained for the task using masked loss. This task utilizes the concept of image-captioning, where an image is provided as an input. The hidden layers of the architecture perform the feature extraction process. Then, these features are passed through the RNN layer coupled with the attention layer to generate a radiology report~\cite{image_captioning}. The image segmentation model uses the same model as the abnormality detection model with the last layers replaced and modified to predict the segmentation task, trained using DICE loss~\cite{image_segmentation}.

    \item The Second step involves performing knowledge distillation between a shallow CNN neural network which will be the backbone of this model. This neural network consists of different prediction heads to perform multiple tasks. The extracted feature representation of the image from the model is passed into the image classification head, text generation head, and image segmentation head to perform abnormality classification, report generation, and image segmentation, simultaneously. Knowledge distillation is performed between the backbone including the specific task head and the individual model trained for the specific task. while the other prediction heads are frozen (C.f.,~Algorithm:~\ref{Student_model}). This is performed until all the prediction heads have acquired knowledge from the complex neural network through knowledge distillation. Here the shallow CNN neural network acts like a student and the individual complex models as teachers. 

    \item The Third and last step involves generating interpretations from the new model which has acquired knowledge from a complex deep neural network model. This step uses model-agnostic and model-specific interpretation techniques to enable interpretability. The LIME interpretations and GradCAM visualizations are incorporated to interpret the abnormality classification model and chest X-ray segmentation model. For report generation, attention visualizer~\cite{attention_viz} is used to output the results. 
\end{enumerate}

\section{Results}
\begin{figure}
        \centering
    \includegraphics[scale=0.7]{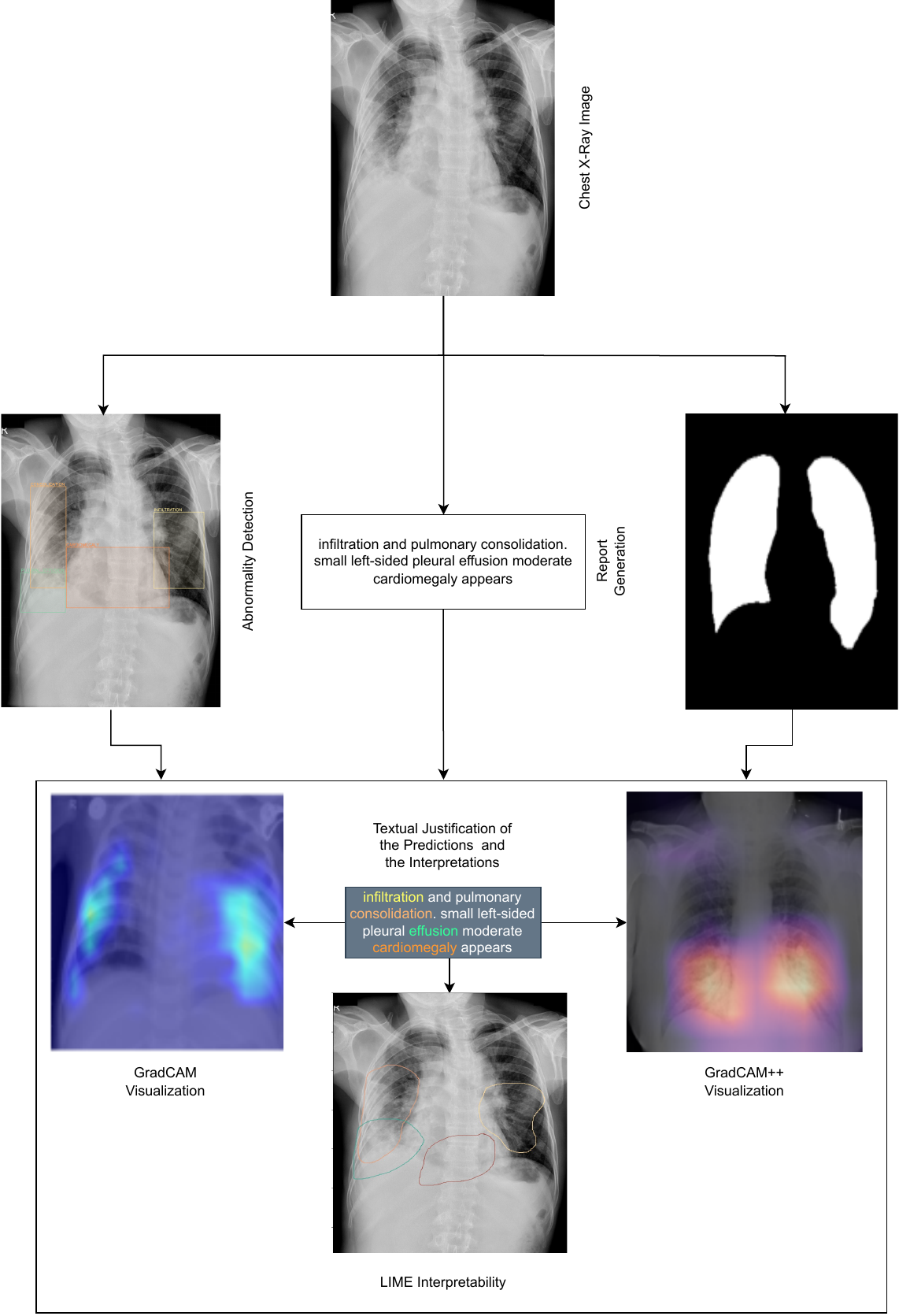}
    \caption{Overview of the Storytelling XAI framework applied in the medical imaging domain. 
    The input image is provided as an input to the student model. 
    The student model performs abnormality detection, lung segmentation, and report text generation. 
    The individual results are provided as input to the interpretability module to generate interpretations. 
    The attention map visualization shows different detected abnormalities.}
    \label{fig:result}
\end{figure}
The chest X-ray image was given as input to the shallow neural network, which was trained with the knowledge distillation method from three different deep neural networks for a different task. The original image was resized to a resolution of 512x512 pixels. See figure:~\ref{fig:result}. The model accurately segmented the two lungs. The abnormality detection model was able to localize and classify the localized affected regions into infiltration, consolidation, pleural effusion, and cardiomegaly. The same model generated the resulting text for radiologists and healthcare professionals. Furthermore, interpretability techniques were applied, which include LIME interpretation, attention map visualization of the generated report, and CAM analysis consisting of GradCAM and GradCAM++. As shown in the figure, the text justification asserts the output of interpretability techniques along with the prediction from the student model which will be beneficial for ML practitioners and at the same time it is able to assert the abnormality detection for healthcare professionals as well. Knowledge distillation is a well-capable method to let the shallow neural network to learn representation over a variety of tasks enabling an interpretable neural network. The benefit of the multiple tasks could be inferred from the result (Fig:\ref{fig:result}), as these multiple tasks when combined, improve the overall understanding of the predictions.
The report generation task serves as a textual justification for the other two tasks, namely abnormality detection and chest X-ray segmentation.
For interpretation, we have focused on LIME, GradCAM, and visualization techniques.
The framework could be extended to incorporate other interpretability techniques including but not limited to SHapley Additive exPlanations (SHAP)~\cite{shap}, Layer-wise Relevance Propagation (LRP)~\cite{lrp}, etc.

\section{Conclusion}
Our framework is a step toward enabling audience-centric explainability. 
We have used concept-based explainability to explain the results of the AI models to healthcare professionals. 
This framework serves as a baseline to demonstrate that the XAI is an amalgamation of interpretability and concept-based approaches to achieve end-to-end explainability.
Note that ML practitioners can also interpret the results of the models in technical terms. 
This framework is also able to explain complex deep neural networks because of the knowledge distillation. 
The teacher models can each be a very complex deep neural network, but the student model can learn the knowledge. 
This framework requires multiple datasets for multiple machine tasks.
The multiple data sets from different sources result in more generalized and robust models in a real-world scenario.
Robust and generalized models that perform well in a variety of real-world scenarios are more likely to be trusted and understood by users.
Robust and generalized models are less susceptible to adversarial manipulation, improving the reliability of their explanations.

In this research, we have limited our investigation to some specific tasks of abnormality detection, report generation, and chest X-ray segmentation.
Other tasks could be added if the datasets for these tasks are available and the inputs are common to all tasks.
We have demonstrated a healthcare use case, but this framework could be extended to other domains given the available datasets.
Multi-task models are a great way to exploit the relationships between different commonalities. 
This technique allows the model to learn shared feature representations. 
By leveraging the shared representations across related tasks, multi-task learning can learn more efficiently from limited data compared to
Learning a shared representation that performs well across multiple related tasks can lead to more stable and meaningful predictions.
Our Storytelling XAI framework can be implemented in critical-stakes domains to enable audience-centric explanations with robust predictions and generalized models, without compromising the performance of even the most complex deep neural networks.

\section{Discussions}

A potential concern is that providing an excessive number of explanations or information may lead to an overload of data, hindering comprehension and user satisfaction. However, the presentation of multiple explanations can enhance transparency and facilitate trust in AI models by addressing the diverse needs of users. By offering a range of explanations, users can gain a more nuanced understanding of the decision-making process, which can ultimately enhance their satisfaction and comprehension~\cite{yang2023survey}.
One might also inquire as to the rationale behind the use of shallow models. The use of shallow models is justified by their reduced susceptibility to overfitting and superior generalization capabilities when processing new data sets, particularly in the context of limited data. Training shallow models for multi-tasking allows for the exploitation of relationships between different input modalities, which is not feasible for single-task models. The majority of interpretability methods employ approximation techniques. As the depth of a neural network model increases, the reliance on approximation increases and reliability decreases. Consequently, shallow methods align well with the aforementioned criteria~\cite{xu2022comparative}. 
\section{Limitations}

Our framework currently lacks the ability of a user to interact with the XAI results freely. 
This may be resolved with a user interface that focuses on integrating Attention Maps, LIME, SHAP, and GradCam visualizations.
When extending this framework to other domains, it may be somewhat problematic to collect different datasets for different tasks. 
There is limited support for teacher models.
The current possibilities are to pick either CNN or RNN-based architectures.
For transformer architectures, it may be difficult to perform knowledge distillation.
The process of knowledge distillation of transformer models is inherently challenging due to the intricate architectural complexity, the considerable number of parameters, and the difficulty in transferring the intricate patterns learned by self-attention mechanisms. Furthermore, the process is further complicated by the necessity to balance multiple objectives and the risk of overfitting.
We have experimented with tasks where the input is solely an image. 
There is indeed not only a limitation in terms of multimodal support but also textual-only data.
Despite the limitations, the work lays down the foundation for future work.

\section{Future Work}

In future work, we would develop a user interface with a human-in-the-loop, preferably domain experts which satisfies the nested model for AI design and validation~\cite{dubey2024nested}. 
Then we would focus on conducting a design study and include a new metric to calculate an agreement score between the framework and the domain experts.
There would also be an extensive effort to integrate the XAI Question Bank (XAI-QB) into our framework~\cite{liao2020questioning}. 
The XAI-QB is critical to the development of explainable AI systems and is designed to help XAI practitioners understand the explanation needs of different end users. 
It represents common user expectations and covers major categories of questions for standard supervised ML systems. 
\clearpage
\bibliography{mybibfile}

\begin{thebibliography}{38}
\expandafter\ifx\csname natexlab\endcsname\relax\def\natexlab#1{#1}\fi
\providecommand{\url}[1]{\texttt{#1}}
\providecommand{\href}[2]{#2}
\providecommand{\path}[1]{#1}
\providecommand{\DOIprefix}{doi:}
\providecommand{\ArXivprefix}{arXiv:}
\providecommand{\URLprefix}{URL: }
\providecommand{\Pubmedprefix}{pmid:}
\providecommand{\doi}[1]{\href{http://dx.doi.org/#1}{\path{#1}}}
\providecommand{\Pubmed}[1]{\href{pmid:#1}{\path{#1}}}
\providecommand{\bibinfo}[2]{#2}
\ifx\xfnm\relax \def\xfnm[#1]{\unskip,\space#1}\fi
\bibitem[{Celi et~al.(2019)Celi, Fine, and Stone}]{celi2019awakening}
\bibinfo{author}{L.~A. Celi}, \bibinfo{author}{B.~Fine}, \bibinfo{author}{D.~J. Stone},
\newblock \bibinfo{title}{An awakening in medicine: the partnership of humanity and intelligent machines},
\newblock \bibinfo{journal}{The Lancet Digital Health} \bibinfo{volume}{1} (\bibinfo{year}{2019}) \bibinfo{pages}{e255--e257}.
\bibitem[{Davenport and Glaser(2022)}]{davenport2022factors}
\bibinfo{author}{T.~H. Davenport}, \bibinfo{author}{J.~P. Glaser},
\newblock \bibinfo{title}{Factors governing the adoption of artificial intelligence in healthcare providers},
\newblock \bibinfo{journal}{Discover Health Systems} \bibinfo{volume}{1} (\bibinfo{year}{2022}) \bibinfo{pages}{4}.
\bibitem[{Reddy(2022)}]{reddy2022explainability}
\bibinfo{author}{S.~Reddy},
\newblock \bibinfo{title}{Explainability and artificial intelligence in medicine},
\newblock \bibinfo{journal}{The Lancet Digital Health} \bibinfo{volume}{4} (\bibinfo{year}{2022}) \bibinfo{pages}{e214--e215}.
\bibitem[{Amann et~al.(2020)Amann, Blasimme, Vayena, Frey, Madai, and Consortium}]{amann2020explainability}
\bibinfo{author}{J.~Amann}, \bibinfo{author}{A.~Blasimme}, \bibinfo{author}{E.~Vayena}, \bibinfo{author}{D.~Frey}, \bibinfo{author}{V.~I. Madai}, \bibinfo{author}{P.~Consortium},
\newblock \bibinfo{title}{Explainability for artificial intelligence in healthcare: a multidisciplinary perspective},
\newblock \bibinfo{journal}{BMC medical informatics and decision making} \bibinfo{volume}{20} (\bibinfo{year}{2020}) \bibinfo{pages}{1--9}.
\bibitem[{Yang et~al.(2024)Yang, Dai, Dubey, Hirche, and Hattab}]{yang2024whom}
\bibinfo{author}{Z.~Yang}, \bibinfo{author}{X.~Dai}, \bibinfo{author}{A.~Dubey}, \bibinfo{author}{S.~Hirche}, \bibinfo{author}{G.~Hattab},
\newblock \bibinfo{title}{Whom to trust? elective learning for distributed gaussian process regression},
\newblock in: \bibinfo{booktitle}{Proceedings of the 23rd International Conference on Autonomous Agents and Multiagent Systems}, \bibinfo{year}{2024}, pp. \bibinfo{pages}{2020--2028}.
\bibitem[{Jin et~al.(2023)Jin, Fan, Gromala, Pasquier, and Hamarneh}]{jin2023invisible}
\bibinfo{author}{W.~Jin}, \bibinfo{author}{J.~Fan}, \bibinfo{author}{D.~Gromala}, \bibinfo{author}{P.~Pasquier}, \bibinfo{author}{G.~Hamarneh},
\newblock \bibinfo{title}{Invisible users: Uncovering end-users' requirements for explainable ai via explanation forms and goals},
\newblock \bibinfo{journal}{arXiv preprint arXiv:2302.06609}  (\bibinfo{year}{2023}).
\bibitem[{Panwar et~al.(2020)Panwar, Gupta, Siddiqui, Morales-Menendez, Bhardwaj, and Singh}]{example_1}
\bibinfo{author}{H.~Panwar}, \bibinfo{author}{P.~Gupta}, \bibinfo{author}{M.~K. Siddiqui}, \bibinfo{author}{R.~Morales-Menendez}, \bibinfo{author}{P.~Bhardwaj}, \bibinfo{author}{V.~Singh},
\newblock \bibinfo{title}{A deep learning and grad-cam based color visualization approach for fast detection of covid-19 cases using chest x-ray and ct-scan images},
\newblock \bibinfo{journal}{Chaos, Solitons \& Fractals} \bibinfo{volume}{140} (\bibinfo{year}{2020}) \bibinfo{pages}{110190}. \URLprefix \url{https://www.sciencedirect.com/science/article/pii/S0960077920305865}. \DOIprefix\doi{https://doi.org/10.1016/j.chaos.2020.110190}.
\bibitem[{prasad Koyyada and Singh(2023)}]{example_2}
\bibinfo{author}{S.~prasad Koyyada}, \bibinfo{author}{T.~P. Singh},
\newblock \bibinfo{title}{An explainable artificial intelligence model for identifying local indicators and detecting lung disease from chest x-ray images},
\newblock \bibinfo{journal}{Healthcare Analytics} \bibinfo{volume}{4} (\bibinfo{year}{2023}) \bibinfo{pages}{100206}. \URLprefix \url{https://www.sciencedirect.com/science/article/pii/S2772442523000734}. \DOIprefix\doi{https://doi.org/10.1016/j.health.2023.100206}.
\bibitem[{Martens et~al.(2023)Martens, Dams, Hinns, and Vergouwen}]{tell_me_a_story}
\bibinfo{author}{D.~Martens}, \bibinfo{author}{C.~Dams}, \bibinfo{author}{J.~Hinns}, \bibinfo{author}{M.~Vergouwen},
\newblock \bibinfo{title}{Tell me a story! narrative-driven xai with large language models},
\newblock \bibinfo{journal}{arXiv preprint arXiv:2309.17057}  (\bibinfo{year}{2023}).
\bibitem[{Asemota and Hooker(2024)}]{asemota2024longitudinal}
\bibinfo{author}{A.~Asemota}, \bibinfo{author}{G.~Hooker},
\newblock \bibinfo{title}{Longitudinal counterfactuals: Constraints and opportunities},
\newblock \bibinfo{journal}{arXiv preprint arXiv:2403.00105}  (\bibinfo{year}{2024}).
\bibitem[{Baron(2023)}]{baron2023explainable}
\bibinfo{author}{S.~Baron},
\newblock \bibinfo{title}{Explainable ai and causal understanding: Counterfactual approaches considered},
\newblock \bibinfo{journal}{Minds and Machines} \bibinfo{volume}{33} (\bibinfo{year}{2023}) \bibinfo{pages}{347--377}.
\bibitem[{Hartmann et~al.(2022)Hartmann, Du, Feldhus, Kruijff-Korbayov{\'a}, and Sonntag}]{xaines}
\bibinfo{author}{M.~Hartmann}, \bibinfo{author}{H.~Du}, \bibinfo{author}{N.~Feldhus}, \bibinfo{author}{I.~Kruijff-Korbayov{\'a}}, \bibinfo{author}{D.~Sonntag},
\newblock \bibinfo{title}{Xaines: Explaining ai with narratives},
\newblock \bibinfo{journal}{KI-K{\"u}nstliche Intelligenz} \bibinfo{volume}{36} (\bibinfo{year}{2022}) \bibinfo{pages}{287--296}.
\bibitem[{Hinton et~al.(2015)Hinton, Vinyals, and Dean}]{hinton2015distilling}
\bibinfo{author}{G.~Hinton}, \bibinfo{author}{O.~Vinyals}, \bibinfo{author}{J.~Dean},
\newblock \bibinfo{title}{Distilling the knowledge in a neural network},
\newblock \bibinfo{journal}{arXiv preprint arXiv:1503.02531}  (\bibinfo{year}{2015}).
\bibitem[{Carrillo et~al.(2021)Carrillo, Cant{\'u}, and Noriega}]{carrillo2021individual}
\bibinfo{author}{A.~Carrillo}, \bibinfo{author}{L.~F. Cant{\'u}}, \bibinfo{author}{A.~Noriega},
\newblock \bibinfo{title}{Individual explanations in machine learning models: A survey for practitioners},
\newblock \bibinfo{journal}{arXiv preprint arXiv:2104.04144}  (\bibinfo{year}{2021}).
\bibitem[{Han et~al.(2023)Han, Kim, Choi, and Yoon}]{han2023impact}
\bibinfo{author}{H.~Han}, \bibinfo{author}{S.~Kim}, \bibinfo{author}{H.-S. Choi}, \bibinfo{author}{S.~Yoon},
\newblock \bibinfo{title}{On the impact of knowledge distillation for model interpretability},
\newblock \bibinfo{journal}{arXiv preprint arXiv:2305.15734}  (\bibinfo{year}{2023}).
\bibitem[{Liu et~al.(2018)Liu, Wang, and Matwin}]{liu2018improving}
\bibinfo{author}{X.~Liu}, \bibinfo{author}{X.~Wang}, \bibinfo{author}{S.~Matwin},
\newblock \bibinfo{title}{Improving the interpretability of deep neural networks with knowledge distillation},
\newblock in: \bibinfo{booktitle}{2018 IEEE International Conference on Data Mining Workshops (ICDMW)}, \bibinfo{organization}{IEEE}, \bibinfo{year}{2018}, pp. \bibinfo{pages}{905--912}.
\bibitem[{Crawshaw(2020)}]{crawshaw2020multi}
\bibinfo{author}{M.~Crawshaw},
\newblock \bibinfo{title}{Multi-task learning with deep neural networks: A survey},
\newblock \bibinfo{journal}{arXiv preprint arXiv:2009.09796}  (\bibinfo{year}{2020}).
\bibitem[{Ribeiro et~al.(2016)Ribeiro, Singh, and Guestrin}]{ribeiro2016should}
\bibinfo{author}{M.~T. Ribeiro}, \bibinfo{author}{S.~Singh}, \bibinfo{author}{C.~Guestrin},
\newblock \bibinfo{title}{" why should i trust you?" explaining the predictions of any classifier},
\newblock in: \bibinfo{booktitle}{Proceedings of the 22nd ACM SIGKDD international conference on knowledge discovery and data mining}, \bibinfo{year}{2016}, pp. \bibinfo{pages}{1135--1144}.
\bibitem[{Linardatos et~al.(2020)Linardatos, Papastefanopoulos, and Kotsiantis}]{linardatos2020explainable}
\bibinfo{author}{P.~Linardatos}, \bibinfo{author}{V.~Papastefanopoulos}, \bibinfo{author}{S.~Kotsiantis},
\newblock \bibinfo{title}{Explainable ai: A review of machine learning interpretability methods},
\newblock \bibinfo{journal}{Entropy} \bibinfo{volume}{23} (\bibinfo{year}{2020}) \bibinfo{pages}{18}.
\bibitem[{Hadji~Misheva et~al.(2021)Hadji~Misheva, Jaggi, Posth, Gramespacher, and Osterrieder}]{hadji2021audience}
\bibinfo{author}{B.~Hadji~Misheva}, \bibinfo{author}{D.~Jaggi}, \bibinfo{author}{J.-A. Posth}, \bibinfo{author}{T.~Gramespacher}, \bibinfo{author}{J.~Osterrieder},
\newblock \bibinfo{title}{Audience-dependent explanations for ai-based risk management tools: A survey},
\newblock \bibinfo{journal}{Frontiers in Artificial Intelligence} \bibinfo{volume}{4} (\bibinfo{year}{2021}) \bibinfo{pages}{794996}.
\bibitem[{Prentzas et~al.(2023)Prentzas, Kakas, and Pattichis}]{prentzas2023explainable}
\bibinfo{author}{N.~Prentzas}, \bibinfo{author}{A.~Kakas}, \bibinfo{author}{C.~S. Pattichis},
\newblock \bibinfo{title}{Explainable ai applications in the medical domain: a systematic review},
\newblock \bibinfo{journal}{arXiv preprint arXiv:2308.05411}  (\bibinfo{year}{2023}).
\bibitem[{Rong et~al.(2023)Rong, Leemann, Nguyen, Fiedler, Qian, Unhelkar, Seidel, Kasneci, and Kasneci}]{rong2023towards}
\bibinfo{author}{Y.~Rong}, \bibinfo{author}{T.~Leemann}, \bibinfo{author}{T.-T. Nguyen}, \bibinfo{author}{L.~Fiedler}, \bibinfo{author}{P.~Qian}, \bibinfo{author}{V.~Unhelkar}, \bibinfo{author}{T.~Seidel}, \bibinfo{author}{G.~Kasneci}, \bibinfo{author}{E.~Kasneci},
\newblock \bibinfo{title}{Towards human-centered explainable ai: A survey of user studies for model explanations},
\newblock \bibinfo{journal}{IEEE Transactions on Pattern Analysis and Machine Intelligence}  (\bibinfo{year}{2023}).
\bibitem[{Johnson et~al.(2019)Johnson, Lungren, Peng, Lu, Mark, Berkowitz, and Horng}]{johnson2019mimic}
\bibinfo{author}{A.~Johnson}, \bibinfo{author}{M.~Lungren}, \bibinfo{author}{Y.~Peng}, \bibinfo{author}{Z.~Lu}, \bibinfo{author}{R.~Mark}, \bibinfo{author}{S.~Berkowitz}, \bibinfo{author}{S.~Horng},
\newblock \bibinfo{title}{Mimic-cxr-jpg-chest radiographs with structured labels},
\newblock \bibinfo{journal}{PhysioNet}  (\bibinfo{year}{2019}).
\bibitem[{Demner-Fushman et~al.(2016)Demner-Fushman, Kohli, Rosenman, Shooshan, Rodriguez, Antani, Thoma, and McDonald}]{demner2016preparing}
\bibinfo{author}{D.~Demner-Fushman}, \bibinfo{author}{M.~D. Kohli}, \bibinfo{author}{M.~B. Rosenman}, \bibinfo{author}{S.~E. Shooshan}, \bibinfo{author}{L.~Rodriguez}, \bibinfo{author}{S.~Antani}, \bibinfo{author}{G.~R. Thoma}, \bibinfo{author}{C.~J. McDonald},
\newblock \bibinfo{title}{Preparing a collection of radiology examinations for distribution and retrieval},
\newblock \bibinfo{journal}{Journal of the American Medical Informatics Association} \bibinfo{volume}{23} (\bibinfo{year}{2016}) \bibinfo{pages}{304--310}.
\bibitem[{Moon et~al.(2022)Moon, Lee, Shin, Kim, and Choi}]{moon2022multi}
\bibinfo{author}{J.~H. Moon}, \bibinfo{author}{H.~Lee}, \bibinfo{author}{W.~Shin}, \bibinfo{author}{Y.-H. Kim}, \bibinfo{author}{E.~Choi},
\newblock \bibinfo{title}{Multi-modal understanding and generation for medical images and text via vision-language pre-training},
\newblock \bibinfo{journal}{IEEE Journal of Biomedical and Health Informatics} \bibinfo{volume}{26} (\bibinfo{year}{2022}) \bibinfo{pages}{6070--6080}.
\bibitem[{Chizhikova et~al.(2023)Chizhikova, Diaz-Galiano, Lopez, and Mart{\'\i}n-Valdivia}]{chizhikova2023sinai}
\bibinfo{author}{M.~Chizhikova}, \bibinfo{author}{M.~Diaz-Galiano}, \bibinfo{author}{L.~A.~U. Lopez}, \bibinfo{author}{M.~T. Mart{\'\i}n-Valdivia},
\newblock \bibinfo{title}{Sinai at radsum23: Radiology report summarization based on domain-specific sequence-to-sequence transformer model},
\newblock in: \bibinfo{booktitle}{The 22nd Workshop on Biomedical Natural Language Processing and BioNLP Shared Tasks}, \bibinfo{year}{2023}, pp. \bibinfo{pages}{530--534}.
\bibitem[{DungNB et~al.(2020)DungNB, Nguyen, Elliott, KeepLearning, NguyenThanhNhan, and Culliton}]{vinbigdata-chest-xray-abnormalities-detection}
\bibinfo{author}{DungNB}, \bibinfo{author}{H.~Q. Nguyen}, \bibinfo{author}{J.~Elliott}, \bibinfo{author}{KeepLearning}, \bibinfo{author}{NguyenThanhNhan}, \bibinfo{author}{P.~Culliton}, \bibinfo{title}{Vinbigdata chest x-ray abnormalities detection}, \bibinfo{year}{2020}. \URLprefix \url{https://kaggle.com/competitions/vinbigdata-chest-xray-abnormalities-detection}.
\bibitem[{Nguyen et~al.(2012)}]{nguyen2012vindr}
\bibinfo{author}{H.~Nguyen}, et~al.,
\newblock \bibinfo{title}{Vindr-cxr: An open dataset of chest x-rays with radiologist’s annotations.(2022) doi: 10.48550},
\newblock \bibinfo{journal}{arXiv}  (\bibinfo{year}{2012}).
\bibitem[{Shivadekar et~al.(2023)Shivadekar, Kataria, Hundekari, Wanjale, Balpande, and Suryawanshi}]{resnet50}
\bibinfo{author}{S.~Shivadekar}, \bibinfo{author}{B.~Kataria}, \bibinfo{author}{S.~Hundekari}, \bibinfo{author}{K.~Wanjale}, \bibinfo{author}{V.~P. Balpande}, \bibinfo{author}{R.~Suryawanshi},
\newblock \bibinfo{title}{Deep learning based image classification of lungs radiography for detecting covid-19 using a deep cnn and resnet 50},
\newblock \bibinfo{journal}{International Journal of Intelligent Systems and Applications in Engineering} \bibinfo{volume}{11} (\bibinfo{year}{2023}) \bibinfo{pages}{241--250}.
\bibitem[{Aneja et~al.(2018)Aneja, Deshpande, and Schwing}]{image_captioning}
\bibinfo{author}{J.~Aneja}, \bibinfo{author}{A.~Deshpande}, \bibinfo{author}{A.~G. Schwing},
\newblock \bibinfo{title}{Convolutional image captioning},
\newblock in: \bibinfo{booktitle}{Proceedings of the IEEE conference on computer vision and pattern recognition}, \bibinfo{year}{2018}, pp. \bibinfo{pages}{5561--5570}.
\bibitem[{Rashid et~al.(2018)Rashid, Akram, and Hassan}]{image_segmentation}
\bibinfo{author}{R.~Rashid}, \bibinfo{author}{M.~U. Akram}, \bibinfo{author}{T.~Hassan},
\newblock \bibinfo{title}{Fully convolutional neural network for lungs segmentation from chest x-rays},
\newblock in: \bibinfo{booktitle}{Image Analysis and Recognition: 15th International Conference, ICIAR 2018, P{\'o}voa de Varzim, Portugal, June 27--29, 2018, Proceedings 15}, \bibinfo{organization}{Springer}, \bibinfo{year}{2018}, pp. \bibinfo{pages}{71--80}.
\bibitem[{Liu et~al.(2021)Liu, Yin, Wu, Ge, Zou, Zhang, and Sun}]{attention_viz}
\bibinfo{author}{F.~Liu}, \bibinfo{author}{C.~Yin}, \bibinfo{author}{X.~Wu}, \bibinfo{author}{S.~Ge}, \bibinfo{author}{Y.~Zou}, \bibinfo{author}{P.~Zhang}, \bibinfo{author}{X.~Sun},
\newblock \bibinfo{title}{Contrastive attention for automatic chest x-ray report generation},
\newblock \bibinfo{journal}{arXiv preprint arXiv:2106.06965}  (\bibinfo{year}{2021}).
\bibitem[{Arslan et~al.(2022)Arslan, Lebichot, Allix, Veiber, Lefebvre, Boytsov, Goujon, Bissyand{\'e}, and Klein}]{shap}
\bibinfo{author}{Y.~Arslan}, \bibinfo{author}{B.~Lebichot}, \bibinfo{author}{K.~Allix}, \bibinfo{author}{L.~Veiber}, \bibinfo{author}{C.~Lefebvre}, \bibinfo{author}{A.~Boytsov}, \bibinfo{author}{A.~Goujon}, \bibinfo{author}{T.~F. Bissyand{\'e}}, \bibinfo{author}{J.~Klein},
\newblock \bibinfo{title}{Towards refined classifications driven by shap explanations},
\newblock in: \bibinfo{booktitle}{International Cross-Domain Conference for Machine Learning and Knowledge Extraction}, \bibinfo{organization}{Springer}, \bibinfo{year}{2022}, pp. \bibinfo{pages}{68--81}.
\bibitem[{Montavon et~al.(2019)Montavon, Binder, Lapuschkin, Samek, and M{\"u}ller}]{lrp}
\bibinfo{author}{G.~Montavon}, \bibinfo{author}{A.~Binder}, \bibinfo{author}{S.~Lapuschkin}, \bibinfo{author}{W.~Samek}, \bibinfo{author}{K.-R. M{\"u}ller},
\newblock \bibinfo{title}{Layer-wise relevance propagation: an overview},
\newblock \bibinfo{journal}{Explainable AI: interpreting, explaining and visualizing deep learning}  (\bibinfo{year}{2019}) \bibinfo{pages}{193--209}.
\bibitem[{Yang et~al.(2023)Yang, Wei, Wei, Chen, Huang, Li, Li, Yao, Wang, Gu et~al.}]{yang2023survey}
\bibinfo{author}{W.~Yang}, \bibinfo{author}{Y.~Wei}, \bibinfo{author}{H.~Wei}, \bibinfo{author}{Y.~Chen}, \bibinfo{author}{G.~Huang}, \bibinfo{author}{X.~Li}, \bibinfo{author}{R.~Li}, \bibinfo{author}{N.~Yao}, \bibinfo{author}{X.~Wang}, \bibinfo{author}{X.~Gu}, et~al.,
\newblock \bibinfo{title}{Survey on explainable ai: From approaches, limitations and applications aspects},
\newblock \bibinfo{journal}{Human-Centric Intelligent Systems} \bibinfo{volume}{3} (\bibinfo{year}{2023}) \bibinfo{pages}{161--188}.
\bibitem[{Xu et~al.(2022)Xu, Song, and Hao}]{xu2022comparative}
\bibinfo{author}{S.~Xu}, \bibinfo{author}{Y.~Song}, \bibinfo{author}{X.~Hao},
\newblock \bibinfo{title}{A comparative study of shallow machine learning models and deep learning models for landslide susceptibility assessment based on imbalanced data},
\newblock \bibinfo{journal}{Forests} \bibinfo{volume}{13} (\bibinfo{year}{2022}) \bibinfo{pages}{1908}.
\bibitem[{Dubey et~al.(2024)Dubey, Yang, and Hattab}]{dubey2024nested}
\bibinfo{author}{A.~Dubey}, \bibinfo{author}{Z.~Yang}, \bibinfo{author}{G.~Hattab},
\newblock \bibinfo{title}{A nested model for ai design and validation},
\newblock \bibinfo{journal}{Iscience} \bibinfo{volume}{27} (\bibinfo{year}{2024}).
\bibitem[{Liao et~al.(2020)Liao, Gruen, and Miller}]{liao2020questioning}
\bibinfo{author}{Q.~V. Liao}, \bibinfo{author}{D.~Gruen}, \bibinfo{author}{S.~Miller},
\newblock \bibinfo{title}{Questioning the ai: informing design practices for explainable ai user experiences},
\newblock in: \bibinfo{booktitle}{Proceedings of the 2020 CHI conference on human factors in computing systems}, \bibinfo{year}{2020}, pp. \bibinfo{pages}{1--15}.

\end{thebibliography}

\end{document}